# ADAPTIVE FEATURE REPRESENTATION FOR VISUAL TRACKING

*Yuqi Han, Chenwei Deng, Zengshuo Zhang, Jiatong Li, Baojun Zhao*

Beijing Key Laboratory of Embedded Real-time Information Processing Technique,
Beijing Institute of Technology, Beijing 100081, China
{yuqi_han, cwdeng, zhangzengshuo, lijiatong, zbj}@ bit.edu.cn

## ABSTRACT

Robust feature representation plays significant role in visual tracking. However, it remains a challenging issue, since many factors may affect the experimental performance. The existing method which combine different features by setting them equally with the fixed weight could hardly solve the issues, due to the different statistical properties of different features across various of scenarios and attributes. In this paper, by exploiting the internal relationship among these features, we develop a robust method to construct a more stable feature representation. More specifically, we utilize a co-training paradigm to formulate the intrinsic complementary information of multi-feature template into the efficient correlation filter framework. We test our approach on challenging sequences with illumination variation, scale variation, deformation etc. Experimental results demonstrate that the proposed method outperforms state-of-the-art methods favorably.

*Index Terms*— Visual tracking, correlation filter, multi-feature templates, ADMM

## 1. INTRODUCTION

Visual tracking is an important research field in computer vision with various applications such as human computer interaction, medical imaging, video surveillance, self-driving vehicles etc. During the past decades, large progress has been made in visual tracking community and various of algorithms have been proposed [1].

Wang et al. [2] break a modern tracker down into five constituent parts, namely, motion model, feature extractor, observation model, model update, and ensemble post-processor to better understand and diagnose visual tracking system. According to their research, effective feature representation plays significant role in a tracker. Although various of visual features (e.g., color histogram, Haar-like, Histogram of Oriented Gradients (HOG)) are utilized for tracking. These features have demonstrated to be lack of discriminability for the modeling of dynamic object appearances. For example, edge-based feature like HOG is sensitive to the spatial configuration of the target and perform poorly when the target undergoes severe deformation. While color feature is sensitive to illumination variation and could hardly distinguish the target from the background.

To overcome the above issue, one strategy is to combine different complementary features in visual object tracker. In [3], Yang et al. propose to integrate Histogram of Oriented Gradients (HOG) feature and Color-Naming (CN) feature together to improve the tracking performance. In [4], Hong et al. integrate various types of feature into a multi-task sparse learning formulation, with each type of feature considered as an individual task. However, both methods mentioned above have ignored the intrinsic relationship between different features by setting them equally with the fixed weight. Since different features do not perform equally well under the same scene simply concatenating features into a high-dimensional vector may degrade the performance even worse than using single feature.

Seeing from the analysis above, in order to construct a more stable feature representation across different challenging attributes, we jointly consider the underlying relation across different features by making a reasonable assumption that for robust tracking, different features should present similar label values on the same sample. Within the proposed scheme, we incorporate the idea of co-training among the unlabeled samples extracted in tracking process into an efficient correlation filter framework. In addition, to tackle the scale variation in tracking procedure, a simple but effective search strategy is adopted in searching the best scale of the target. Experimental results demonstrate that the proposed method outperforms state-of-the-art methods.

## 2. PROPOSED METHOD

### 2.1 Tracking by correlation framework

Most existing trackers adopt either the generative or the discriminative approach. For generative trackers [5], an elaborated appearance model is often designed to describe a set of target observations in order to find the most proper candidate among numerous observations. On the other hand, the discriminative trackers [6] consider visual tracking as a binary

---
This work was supported by the Chang Jiang Scholars Programme {Grant No. T2012122}, and 111 Project of China under Grant B14010

classification problem which usually employ the trained classifier to distinguish the target from the background and estimate its location in each frame from plenty of candidates. However, the efficiency of such approach is limited by the number of training samples.

To address these issues, Bolme et al. [7] introduce correlation filter into tracking as classifiers. Henriques et al. find that all the translated samples around the target could be collected for training without sacrificing tracking speed with the help of circulant matrix and Ridge Regression [8]. The goal of Ridge Regression training is to learn a filter **w** which could minimize a least-square loss for all the circulant shifts of the training templates x and its label **y**. The training problem could be viewed as minimizing the following function:

$$\min_w \|\mathbf{\Phi w} - \mathbf{y}\|_2^2 + \lambda \|\mathbf{w}\|_2^2 \quad (1)$$

$\Phi$ denotes the mapping of all the circular shifts of the template **x**. If the mapping $\Phi$ is linear, equation (1) could be solved directly using the DFT matrix. Thus, for correlation filter based tracking approach, we could locate the target's position when we obtain its filter **w**.

Several seminal follow-up work have been proposed to enhance the tracking efficiency and robustness, including non-linear kernels and multi-dimensional features [9], scale estimation [3] and context learning [10].

Despite of the huge success of correlation tracking, these trackers could not represent the target robustly across different challenging scenarios since they either use single feature or concatenate multi-features into a high-dimensional feature vector by setting each feature the same weight.

### 2.2 Joint multi-feature integration

In order to address such limitations and construct a more stable feature representation, in this section, we would formulate tracking task into correlation framework and show that how filter **w** could be computed using multiple templates with multiple features. Without loss of generality, our derivation will be presented for 1-D signal, but it could be extended to 2-D signals with multiple templates.

If we regard each feature extraction in equation (1) as one template. We could formulate the training problem as:

$$\min_{w_i} \|\mathbf{\Phi}_i \mathbf{w}_i - \mathbf{y}\|_2^2 + \lambda \|\mathbf{w}_i\|_2^2 \quad (2)$$

$\Phi_i$ is the mapping matrix evaluated for template x and its circulant shift using $i^{th}$ feature, which contains HOG feature, LBP feature, Color-Naming feature etc. While $w_i$ is the filter corresponding to those different features. Ideally, we could extend such equations to **N** different features. It should be mentioned that the feature sizes do not consist with each other at first, and we follow the prior work [11] that alignment should be applied for the feature data for further processing.

In order to explore the internal relation between these different types of features, we make a rational hypothesis that for the same sample, no matter the labeled sample or unlabeled one, different features should give a similar regression label ideally. As for correlation filter based tracking approach, we could obtain serval unlabeled samples from the circulant shift of the target template easily. Afterwards, by incorporating co-training, we could reconstruct the optimal function:

$$\min_w (\sum_{i=1}^{N} \|\mathbf{\Phi}_i \mathbf{w}_i - \mathbf{y}\|_2^2 + \lambda_0 \|\mathbf{w}\|_1 + \sum_{i=1}^{N-1} \sum_{j=i+1}^{N} \lambda_{i,j} \|\mathbf{\Psi}_i \mathbf{w}_i - \mathbf{\Psi}_j \mathbf{w}_j\|_2^2) \quad (3)$$

$$s.t. \quad \mathbf{w} - \sum_{i=1}^{N} \mathbf{G}_i \mathbf{w}_i = \mathbf{0} \quad (4)$$

$\Psi_i$ denotes the mapping matrix evaluated for the unlabeled samples extracted during the tracking procedure using the $i^{th}$ feature. $\lambda_0$ and $\lambda_{i,j}$ are regularization parameter to control the model complexity. The constrain in the above formulation represent that the filter **w** could be represented as: $w = [\mathbf{w}_1; \mathbf{w}_2; \cdots; \mathbf{w}_N]$

The first two term denote the extended version of the Ridge Regression for multi-features. As mentioned, the features' dimensions do not consist with each other, and the training samples always contain some reductant information. Hence, we utilize $L_1$ norm to remove the trivial features and automatically select the important one, as well as aligning the dimension of different features. The last sum term means the output bias predicted by different features on the unlabeled samples during tracking. We assume that it's likely to track the object precisely if different features give similar labels, otherwise the filter for different features would iterate to give a more accurate result.

### 2.3 Optimization algorithm

The solution for optimization model in equation (3) and (4) is not straightforward due to the unsmooth $L_1$ norm. Hence, we would incorporate the alternating direction method of multipliers (ADMM) technique [11] to this optimization. Without loss of generality, the derivation will be highlighted for three filters, but it's appropriate for multiple features similarly. By introducing the Lagrangian multiplier **Y** and a positive penalty parameter $\mu$, we could further reach the following equivalent problem:

$$L(\mathbf{w}_1, \mathbf{w}_2, \mathbf{w}_3, \mathbf{w}, \mathbf{Y}) = \|\mathbf{\Phi}_1 \mathbf{w}_1 - \mathbf{y}\|_2^2 + \|\mathbf{\Phi}_2 \mathbf{w}_2 - \mathbf{y}\|_2^2 + \|\mathbf{\Phi}_3 \mathbf{w}_3 - \mathbf{y}\|_2^2 + \lambda_0 \|\mathbf{w}\|_1 \\ + \lambda_{1,2} \|\mathbf{\Psi}_1 \mathbf{w}_1 - \mathbf{\Psi}_2 \mathbf{w}_2\|_2^2 + \lambda_{2,3} \|\mathbf{\Psi}_2 \mathbf{w}_2 - \mathbf{\Psi}_3 \mathbf{w}_3\|_2^2 + \lambda_{1,3} \|\mathbf{\Psi}_1 \mathbf{w}_1 - \mathbf{\Psi}_3 \mathbf{w}_3\|_2^2 \\ + Tr(\mathbf{Y}^T (\mathbf{w} - \mathbf{w}_1 \mathbf{G}_1 - \mathbf{w}_2 \mathbf{G}_2 - \mathbf{w}_3 \mathbf{G}_3)) + \frac{\mu}{2} \|\mathbf{w} - \mathbf{w}_1 \mathbf{G}_1 - \mathbf{w}_2 \mathbf{G}_2 - \mathbf{w}_3 \mathbf{G}_3\|_2^2 \quad (5)$$

In order to solve the constrained problem, the ADMM technique uses a series of iterations:

$$\begin{cases} \mathbf{w}_1^{(k+1)} = \arg\min L(\mathbf{w}_1, \mathbf{w}_2^{(k)}, \mathbf{w}_3^{(k)}, \mathbf{w}^{(k)}, \mathbf{Y}^{(k)}) \\ \mathbf{w}_2^{(k+1)} = \arg\min L(\mathbf{w}_1^{(k)}, \mathbf{w}_2, \mathbf{w}_3^{(k)}, \mathbf{w}^{(k)}, \mathbf{Y}^{(k)}) \\ \mathbf{w}_3^{(k+1)} = \arg\min L(\mathbf{w}_1^{(k)}, \mathbf{w}_2^{(k)}, \mathbf{w}_3, \mathbf{w}^{(k)}, \mathbf{Y}^{(k)}) \\ \mathbf{w}^{(k+1)} = \arg\min L(\mathbf{w}_1^{(k)}, \mathbf{w}_2^{(k)}, \mathbf{w}_3^{(k)}, \mathbf{w}, \mathbf{Y}^{(k)}) \\ \mathbf{Y}^{(k+1)} = \mathbf{Y}^{(k)} + \mu_k (\mathbf{w}^{(k+1)} - \mathbf{w}_1^{(k+1)} \mathbf{G}_1 - \mathbf{w}_2^{(k+1)} \mathbf{G}_2 - \mathbf{w}_3^{(k+1)} \mathbf{G}_3) \end{cases} \quad (6)$$

The stopping criterion we use in the derivation depends on the residual of the filter in the previous iterations. Once the residual is small enough, the optimization process terminated.

$$\|\mathbf{w}^{(k+1)} - \mathbf{w}_1^{(k+1)}\mathbf{G}_1 - \mathbf{w}_2^{(k+1)}\mathbf{G}_2 - \mathbf{w}_3^{(k+1)}\mathbf{G}_3\| \leq \varepsilon, \text{ where}$$

$\varepsilon$ is the tolerance error. It's worth noting that in the traditional ADMM technique, the value of $\mu$ (shown in equation 6) is increasing in each iteration to guarantee its convergence. As shown in the equation (6), in each iteration, we solve for the filter via alternating fixed-point optimization. Note that the optimization problem above has the same form, we could just derive one of them.

The optimization over $\mathbf{w}_1$ is equivalent to

$$\min_{\mathbf{w}_1} L(\mathbf{w}_1) = \|\Phi_1 \mathbf{w}_1 - \mathbf{y}\|_2^2 + \lambda_{1,2}\|\Psi_1 \mathbf{w}_1 - \Psi_2 \mathbf{w}_2^{(k)}\|_2^2 + \lambda_{1,3}\|\Psi_1 \mathbf{w}_1 - \Psi_3 \mathbf{w}_3^{(k)}\|_2^2 \quad (7)$$
$$-Tr(\mathbf{Y}^T \mathbf{w}_1 \mathbf{G}_1) + \frac{\mu}{2}\|\mathbf{w}_1 \mathbf{G}_1 + (\mathbf{w}_2^{(k)}\mathbf{G}_2 + \mathbf{w}_3^{(k)}\mathbf{G}_3 - \mathbf{w}^{(k)})\|_2^2$$

For the convenience of calculation, we define $\mathbf{A}^{(k)} = (\mathbf{w}_2^{(k)}\mathbf{G}_2 + \mathbf{w}_3^{(k)}\mathbf{G}_3 - \mathbf{w}^{(k)})$. We set the gradient of equation (7) equals to zero, and identify the left hand to the right. By doing so, we obtain

$$(2\Phi_1^T \Phi_1 + 2\lambda_{1,2}\Psi_1^T\Psi_1^T + 2\lambda_{1,3}\Psi_1^T\Psi_1^T + \mu \mathbf{G}_1 \mathbf{G}_1^T) \mathbf{w}_1$$
$$= 2\Phi_1^T \mathbf{y} + 2\lambda_{1,2}\Psi_1^T\Psi_2 \mathbf{w}_2^{(k)} + 2\lambda_{1,3}\Psi_1^T\Psi_3 \mathbf{w}_3^{(k)} + \mathbf{Y}\mathbf{G}_1^T + \mu \mathbf{A}^{(k)}\mathbf{G}_1^T \quad (8)$$

By using the similar derivation, we could yield $\mathbf{B}^{(k)}$ and $\mathbf{C}^{(k)}$ with similar form.

The optimization over $\mathbf{w}$ equals to:

$$\min_{\mathbf{w}} L(\mathbf{w}) = \lambda_0 \|\mathbf{w}\|_1 + \frac{\mu}{2}\|\mathbf{w} - \mathbf{V}\|_2^2 \quad (9)$$

where $\mathbf{V} = \frac{1}{\mu}\mathbf{Y}^{(k)} + \mathbf{w}_1^{(k+1)}\mathbf{G}_1 + \mathbf{w}_2^{(k+1)}\mathbf{G}_2 + \mathbf{w}_3^{(k+1)}\mathbf{G}_3$

Then the optimal solution to $\mathbf{w}$ could be obtained as:

$$\mathbf{w} = \begin{cases} (1 - \frac{\lambda_0}{\mu\|\mathbf{V}\|_2})\mathbf{V} & \|\mathbf{V}\|_2 > \frac{\lambda_0}{\mu} \\ 0 & \text{otherwise} \end{cases} \quad (10)$$

It should be mentioned that for ADMM model with multiple blocks $(\mathbf{w}, \mathbf{w}_1\mathbf{G}_1, \mathbf{w}_2\mathbf{G}_2, \mathbf{w}_3\mathbf{G}_3)$ in the proposed method, its convergence has been well proven in [12]. According to its theoretical derivation, $\{\mu_k\}$ is non-decreasing with $\sum_{k=1}^{\infty} \mu_k^{-1} = +\infty$, which demonstrates that $\mu_k$ wouldn't grow too fast (k means the $k^{th}$ iteration).

## 2.4 Search Strategy

During the tracking procedure, the target may undergo scale variations. We incorporate a simple yet efficient multi-scale searching scheme to counteract this issue. Rather than search the translation and scale variation jointly, we set a series of scale sizes to estimate the best scale of the target after obtaining the target position predicted by the filter. Note that for translation and scale, we only search the region near the target instead of the whole image. In contrast of the previous work [3] that consider all the scale size equally, we make a reasonable assumption that the prior probability of the scales set follows Gaussian distribution since the scale of target doesn't change significantly between consecutive frames. The experiment shows that such search strategy gives a more accurate results compared to the scheme adopted in the previous work.

## 3. EXPERIMENTS

To evaluate the effectiveness and robustness of the tracker, we empirically validated the proposed tracker on 6 challenging sequences. Furthermore, we compare our tracker with seven state-of-the-art methods. These trackers could be broadly categorized into three classes: (i) baseline correlation filter based trackers including CSK [8], KCF [9] and STC [10], (ii) trackers using multiple features such as SAMF [3] and MTMV [4], (iii) other representative trackers reported in [2] such as SCM [5] and Struck [6] methods.

### 3.1. Quantitative results

We evaluated all the trackers by adopting one common criteria: the overlap ratio. We denote the ratio S = Area (BT∩BG)/ Area (BT∪BG), where BT is the tracked bounding box and BG denotes the ground truth. The overlap ratio shows the percentage of frames with S > t, throughout all threshold t ∈ [0,1]. The average overlap rate is shown in Table 1. It demonstrates that our tracker outperforms other state-of-the-art methods in these sequences.

### 3.2. Qualitative Results

Figure 1 shows qualitative results comparing with the other state-of-the-art trackers on challenging sequences. In *couple and boy*, the appearances of the target are changing a lot because of the pose variation and deformation. Both our tracker and SAMF perform well due to the use of complementary features. While in *shaking and singer2*, the illumination varies a lot. SAMF and MTMV tracker drift from the target because of the same weight on HOG and color feature. While our tracker could locate the target accurately because of the utilization of co-training. The results predicted by HOG feature could correct the ones predicted by CN feature, which is insufficient in such scenarios. In *freeman3 and car4*, the target undergoes large scale change. KCF could not adapt to the scale variation due to the fixed size of the training template. The proposed tracker could tail the changing state for the entire sequences, which could be attributed to the effective scale searching strategy.

**Table 1.** Average overlap rate. The red bold fonts and blue italic fonts indicate the best and the second best performance.

| Sequences | CSK | KCF | STC | SAMF | MTMV | Struck | SCM | Ours |
|---|---|---|---|---|---|---|---|---|
| Boy | 0.646 | 0.762 | 0.536 | 0.74 | *0.767* | 0.747 | 0.37 | **0.78** |
| Car4 | 0.468 | 0.485 | 0.358 | 0.737 | 0.158 | 0.49 | *0.745* | **0.775** |
| Couple | 0.074 | 0.198 | 0.073 | 0.407 | 0.456 | *0.532* | 0.098 | **0.634** |
| Freeman3 | 0.3 | 0.331 | 0.257 | 0.302 | 0.295 | 0.264 | **0.709** | *0.483* |
| Shaking | 0.572 | 0.042 | 0.617 | 0.136 | **0.712** | 0.356 | 0.38 | *0.621* |
| Singer2 | 0.046 | **0.721** | 0.406 | 0.042 | *0.698* | 0.043 | 0.17 | 0.598 |

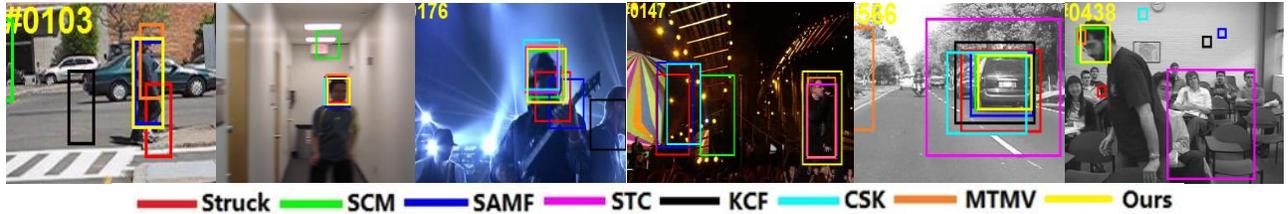

**Fig. 1.** Representative tracking results on challenging sequences

## 4. CONCLUSIONS

In this paper, we propose a joint multi-template feature learning approach for correlation tracking. We explore the intrinsic relationship shared among different features by incorporating the idea of co-training. We show that the optimization problem could be solved effectively by ADMM technique. In the experimental section, we implement our tracking framework using three complementary features, i.e. Hog, color-naming and LBP. But it's worth noting that the proposed method could be extended to other features even from sensors other than optical cameras. Exhaustive experiments demonstrate that the proposed method could outperform the state-of-the-art method on challenging videos.